\documentclass[a4paper,conference]{IEEEtran}
\ifCLASSINFOpdf
  \usepackage[pdftex]{graphicx}
\else
\fi
\usepackage{amsmath}
\usepackage{amsfonts}
\usepackage{mathtools,xparse}
\usepackage[section]{placeins}
\usepackage{float}

\usepackage{listings}
\DeclarePairedDelimiter{\norm}{\lVert}{\rVert}
\NewDocumentCommand{\normL}{ s O{} m }{%
  \IfBooleanTF{#1}{\norm*{#3}}{\norm[#2]{#3}}_{L_2(\Omega)}%
}
\usepackage{hyperref}

\usepackage{amsmath}
\usepackage{algorithm}
\usepackage[noend]{algpseudocode}

\makeatletter
\def\BState{\State\hskip-\ALG@thistlm}
\makeatother
\usepackage{makecell}

\usepackage[nocompress]{cite}

\begin{document}
%
\title{Table Detection in the Wild: A Novel Diverse Table Detection Dataset and Method}

\author{\IEEEauthorblockN{Mrinal Haloi, Shashank Shekhar, Nikhil Fande, Siddhant Swaroop Dash, Sanjay G}
\IEEEauthorblockA{Subex AI Labs\\
Email: mrinalhaloi11@gmail.com, shashekharank@gmail.com}
}


%


\maketitle

\begin{abstract}
Recent deep learning approaches in table detection achieved outstanding performance and proved to be effective in identifying document layouts. Currently, available table detection benchmarks have many limitations, including the lack of samples diversity, simple table structure, the lack of training cases, and samples quality. In this paper, we introduce a diverse large-scale dataset for table detection with more than seven thousand samples containing a wide variety of table structures collected from many diverse sources. In addition to that, we also present baseline results using a convolutional neural network-based method to detect table structure in documents. Experimental results show the superiority of applying convolutional deep learning methods over classical computer vision-based methods. The introduction of this diverse table detection dataset will enable the community to develop high throughput deep learning methods for understanding document layout and tabular data processing. \\
{\normalfont Dataset is available at: 1. \url{https://www.kaggle.com/datasets/mrinalim/stdw-dataset} \\ 2. \url{https://huggingface.co/datasets/n3011/STDW}}
\end{abstract}


%
\IEEEpeerreviewmaketitle

\section{Introduction}

Table detection is one of the crucial tasks in document layout analysis and table data understanding. Especially for extracting data from scanned documents table detection plays a significant role. 

The recent development of applying deep learning in computer vision-related tasks enabled the researcher to develop a state-of-the-art document layout analysis system \cite{layout1, layout2}. Deep learning technique such as convolutional neural network is widely used in segmentation \cite{rethink}, classification \cite{clf2, clf1} and object detection task \cite{maskrcnn}. Building a high-performance convolutional model requires a large dataset resembling the target problem. A good dataset has the attributes of good samples diversity, high-quality ground truths, readable samples resolution, and a lot of training cases. As we have observed, all the current table detection datasets have various limiting aspects. 
 
 Firstly, the quality of samples available in these datasets is not adequate, which makes it hard for the trained model to generalize well in the high-quality sample cases. The structural details of tables are important in the detection and low-quality samples lack clarity in capturing minute details of tables structure.  Hence, samples resolution and quality are important in the training detection model.
 
 \begin{figure}[h]
  \centering
      \includegraphics[width=3.1in,height=2.9in]{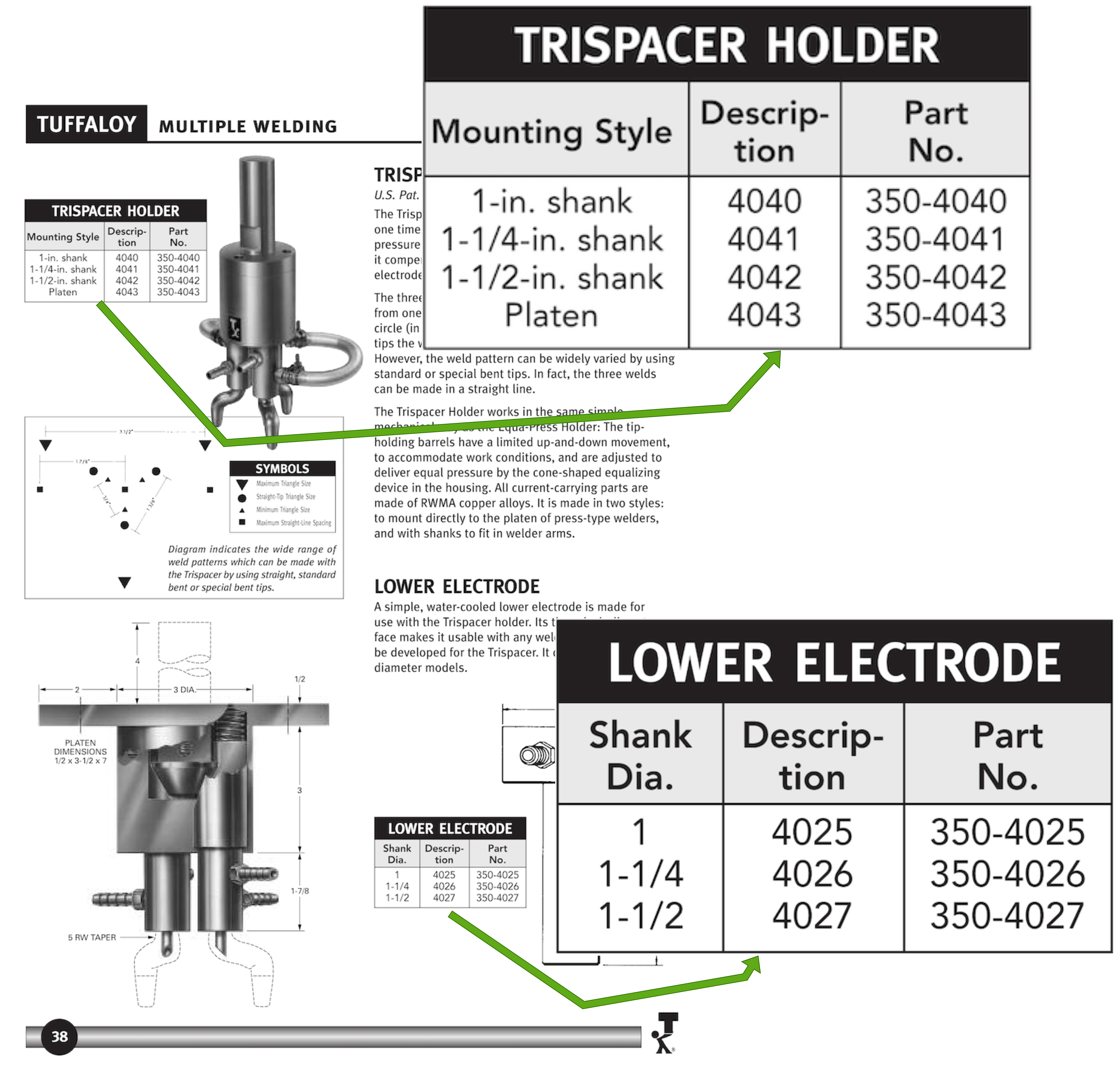}
\caption{Table detection in documents. An example of tables in a datasheet document.}
\label{fig:intro}
\end{figure}
 
 Secondly, the diversity of samples is very limited. Dataset samples should capture all possible variations of table structures starting from a simple table to complex nested table structures including grid and non-grid variations. Deep learning model generalization performance improves with diverse discriminative samples as it learns the underlying pattern present in the dataset.

Finally, the most crucial is the number of available samples in the dataset. Dataset's cardinality has a direct impact on the overall performance. Advanced deep learning model depends on a large dataset, a limited number of samples prevent us from applying complex very deep networks for table detection. A deep learning model built with limited data suffers from overfitting and poor inference time results.
 
To address these problems, we have gathered a new large-scale benchmark dataset for table detection. The presented dataset consists of 7K image samples from diverse sources capturing a wide variety of table occurrences. We have collected table samples from scanned documents, word documents, and searchable pdf documents. Collected samples were checked for quality standards by removing blurry, noisy, and low-quality samples from the final dataset. 

As another contribution, we also provide baseline results using a deep learning-based model and classical selective search method. Additionally, we also provide a novel method to compute the diversity of the samples of a dataset by leveraging latent representation computed using a pre-trained encoder.

The presented dataset can help the researchers to develop a novel table detection method to understand and map document layouts for information extraction. This dataset will make it easy for the community to apply data-dependent algorithms such as deep convolutional neural networks for the task of layout and table detection. 

Our experimental results using the presented deep learning method show promising results in detecting complex table structures. The baseline results were able to outperform the classical computer vision-based methods for detecting tables asserting the advantages of a data-driven deep neural network.  

The rest of the paper is organized as follows: Section 2 explores the current table detection methods and benchmarks. Section 3 introduces the presented dataset, its properties, and evaluation metrics. Section 4 presents the deep convolutional layer-based RetinaNet method and selective search-based detection method for table detection in the wild images. Section 5 shows the experimental results obtained by classical computer vision methods alongside the proposed deep learning method on our benchmark, and section 6 concludes the paper.

\section{Related work}
In this section, we will go through the publicly available table detection datasets and the recent advancements of deep learning in this domain. 

\subsection{Table Detection Datasets}
TableBank \cite{tablebank} dataset is prepared from word and latex documents available on the internet. This dataset contains 417K labeled table images. This dataset does not contain any tables from scanned pdf documents. Another limitation is the diversity of the samples, this dataset was collected from '.docx' format documents and scientific articles from the arxiv.org website.

Marmot \cite{marmot} dataset is consisting of 2000 pdf pages with tables. Most of these examples were collected from research papers.

ICDAR 2013 \cite{icdar} dataset released as part of 2013 competition for both table detection and its structural analysis. It contains 128 samples collected mostly from US and EU government sources. 

UNLV \cite{unlv} dataset contains 427 samples collected from scanned documents. These samples were sourced from magazines, newspapers, corporate reports, business letters, etc. 

DeepFigures \cite{deepfigures} is a large dataset consisting of figures and tables samples. This dataset is collected from latex and XML sources. It has 1.4M induced tables, but it does not contain any scanned documents, also the intra-samples diversity of this dataset is limited.

To summarize our dataset has many advantages over the existing public datasets as follows: 
\begin{itemize}
\item Samples include scanned and searchable documents. 
\item High diversity of samples, capturing a wide range of table designs.
\item High resolution of samples.
\item Biggest dataset with manual annotation.

\end{itemize}

\subsection{Table Detection Methods}
Both classical computer vision and deep learning-based methods are used to solve the table detection problem. Recently convolutional detection-based methods achieved state-of-the-art results on publicly available datasets such as TableBank, ICDAR, Marmot, etc. 

Li et al. \cite{tablebank} proposed a Faster-RCNN \cite{fasterrcnn} based convolutional detection method and reported a performance of 0.93 F1 score in the TableBank dataset. They have used Resnet-101 and Resnet-152 very deep neural networks as feature extractor backbone. 

TableNet \cite{tablenet} is another deep learning method formulate table detection as region segmentation problem and uses FCN architecture \cite{fcn} to segment table region. It uses VGG-16 \cite{vgg} as feature extractor for the FCN model. TableNet has reported a performance of 0.95 F1 score on the iCDAR dataset. 

Another Faster-RCNN based approach proposed by Gilani et al. \cite{gilani} reported state-of-the-art performance on the UNLV dataset.

Schreiber et al. \cite{deepdesrt} uses a deep learning-based method to detect tables and identify table structure by detecting rows, columns, and table cells. It also uses the Faster-RCNN method with VGG-16 as the feature extractor backbone. 

Classical feature engineering approaches were also used to detect tables in scanned and searchable documents. Kasar et al. \cite{kasar} used an SVM classifier on features extracted using horizontal and vertical lines information to predict if a sample has a table or not.  Silva et al. \cite{silva} uses Hidden Markov model to detect tables from documents. Based on the MXY tree data structure Cesarini et al. \cite{cesarini} presented a hierarchical representation for locating tables in document images. 

Despite the successes of applying deep learning models on various datasets, these methods still lack generalization performance on out-of-domain samples due to the constrained nature of the available public datasets. Applying these models to production scenarios is still a work in progress.

\section{The Dataset}
In this section, we introduce the metadata details and the evaluation criteria of STDW table detection dataset. 
Dataset can be accessed in this link: https://github.com/subex/STDW.

\subsection{The STDW Table Detection Dataset}

\begin{figure*}
  \centering
      \includegraphics[width=\textwidth, height=3in]{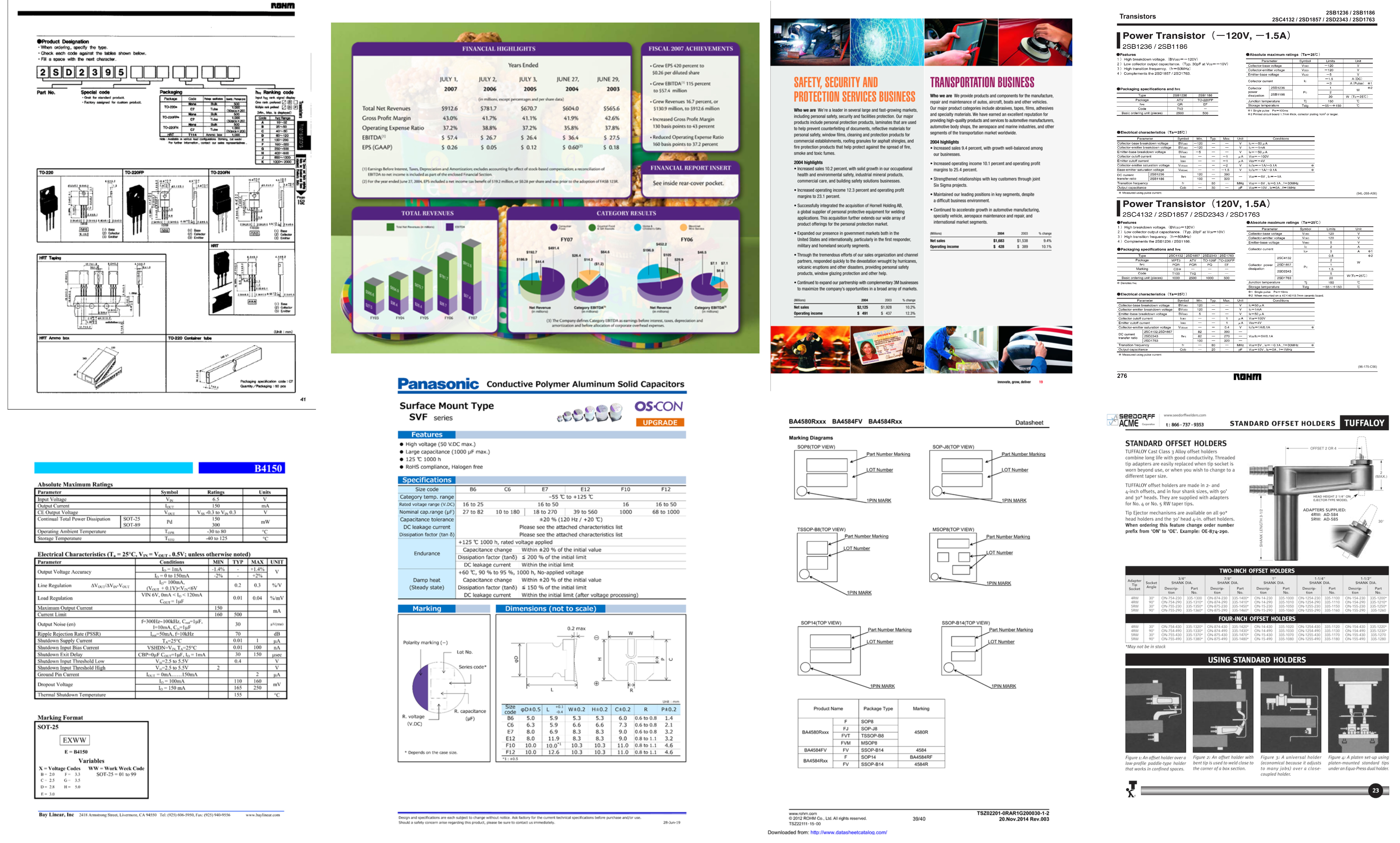}
\caption{Dataset samples images with diverse examples}
\label{fig:sample_images}
\end{figure*}

\textbf{Data Sources:} To prepare the dataset, we have utilized the resources available on the internet. Documents containing tables were collected from various sources to ensure intra-dataset samples diversity. Sources such as electronic component datasheets, material safety data sheets, product safety data sheets, billing invoices, research papers, finance reports, books, etc are used for gathering the samples. Samples contain English, German, Japanese, Hindi, and many other languages capturing diverse scripts. Figure~\ref{fig:sample_images} shows some sample images from the collected dataset.

\textbf{Data Modalities:} Both scanned and searchable documents with tables are included in the dataset. Scanned documents include RGB and grayscale samples. Collected samples have one or more tables in them. Samples resolution various from $500 * 500 * 3$ to $5000 * 5000 * 3$, capturing a wide range of image qualities. 

\begin{figure}
  \centering
      \includegraphics[width=3.3in, height=3.5in]{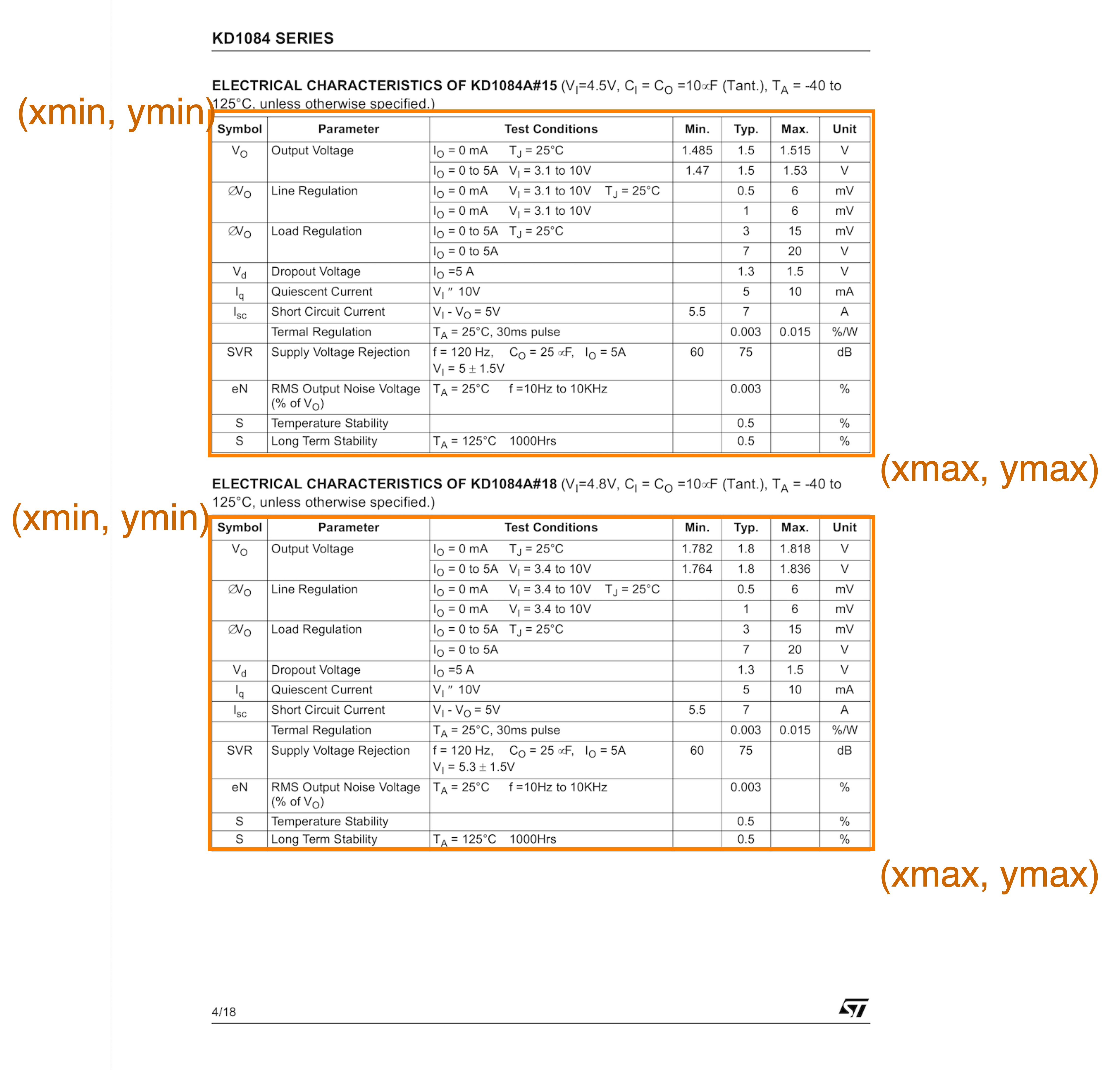}
\caption{An example image with bounding boxes annotation. We denote top-left corner as (xmin, ymin) and bottom-right corner as (xmax, ymax)}
\label{fig:annotation}
\end{figure}

\textbf{Labelling:} All samples in the dataset are labeled manually using the GUI-based Labelme \cite{labelme} annotation tool. Bounding box-based standard detection problem's labeling method is used to label all the images. For each of the image bounding boxes, coordinates are stored in an XML file as shown in the Listing~\ref{lst:annotation} following the PASCAL VOC~\cite{pascal_voc} annotation format.
For each bounding box, top left $(xmin, ymin)$ and bottom-right $(xmax, ymax)$ coordinates are stored. Figure~\ref{fig:annotation} shows a sample image with a bounding box marked on the image showing the position respective coordinates.

\begin{lstlisting}[float,caption=An example XML file content showing the annotation for a single bounding box.,label=lst:annotation]

<annotation>
	<object>
		<name>Table</name>
		<pose>Unspecified</pose>
		<truncated>0</truncated>
		<difficult>0</difficult>
		<bndbox>
			<xmin>541</xmin>
			<ymin>970</ymin>
			<xmax>4060</xmax>
			<ymax>2766</ymax>
		</bndbox>
	</object>
	...
	<object>
	...
	</object>
</annotation>
\end{lstlisting}

\textbf{Data Statistics:}
Table~\ref{table:stdw_stats} depicts the STDW dataset samples statistics.

\begin{figure*}
  \centering
      \includegraphics[width=\textwidth, height=2.5in]{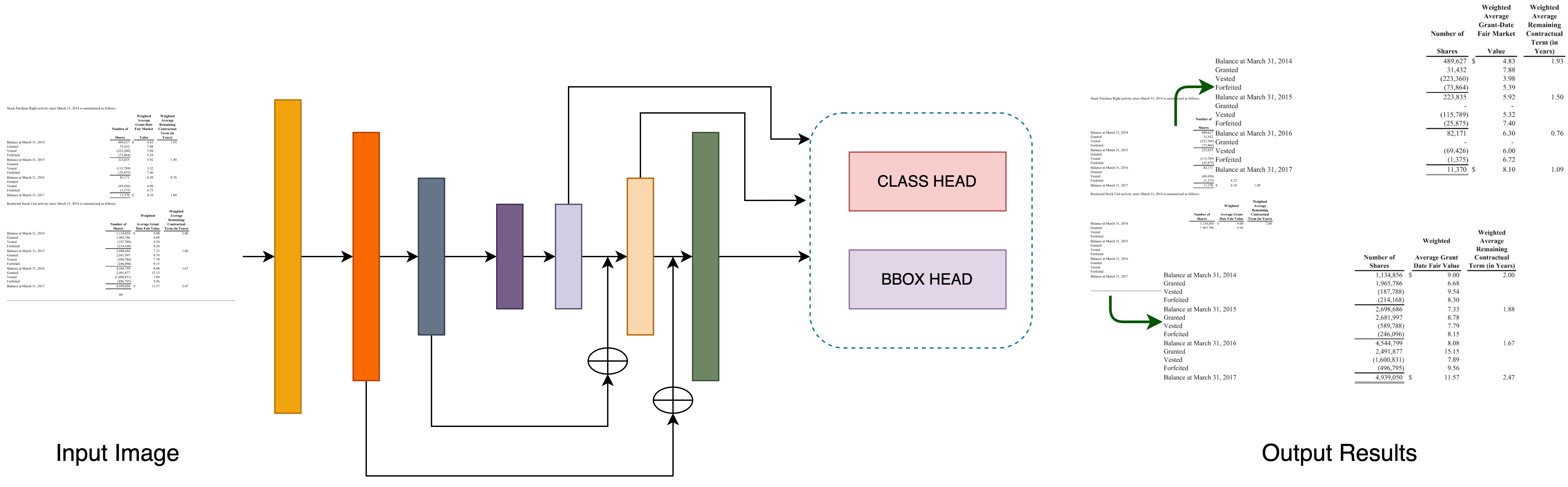}
\caption{RetinaNet based method for Table Detection. RetinaNet uses two task-specific heads, one for predicting objects labels and the other for predicting object bounding boxes.  RetinaNet is a fully convolutional model and it uses upsampling layers and skip connections to build the input layers for the task-specific heads.}
\label{fig:method}
\end{figure*}

\subsection{Benchmark Evaluations}
We define criteria for two types of table detection evaluation metrics to unify reported results on this dataset. Work derived using this benchmark dataset may use these metrics for fair comparisons. We report the performance of the baseline models on
these two metrics. 

\textbf{IoU-Intersection over union:}  Overlap between two bounding boxes is measured using Intersection Over Union (IoU) also known as Jaccard Index. We use this metric to measure the extent of the correctness of the predicted bounding box with the ground truth bounding box. Eq~\ref{eq:1} shows the mathematical formulation on how to calculate IoU between two bounding boxes B1 and B2.

\begin{table}[h]
\begin{center}
  \begin{tabular}{ | p{3cm} | c |c|c| }
    \hline
    \thead{STDW} & \thead{No of Images} & \thead{No of Tables}\\ \hline
    Scanned PDFs & 2345 & 5102 \\ \hline
    Searchable documents & 4945 & 7329 \\
    \hline
  \end{tabular}
\end{center}
\caption{STDW dataset samples statistics. It includes both scanned and searchable/native-digital documents,}
\label{table:stdw_stats}
\end{table}

\begin{equation}
IoU= \frac{|B1 \cap B2|}{|B1 \cup B2|}
\label{eq:1}
\end{equation}

\textbf{AP-Average Precision:} Average Precision (AP) summarizes a precision-recall curve as the weighted mean of the precisions obtained at the different thresholds. The increase in recall from the previous threshold has used a weight. Eq~\ref{eq:2} shows the mathematical formulation on how to compute AP using a precision-recall curve, where $P_{n}$ and $R_{n}$ are the precision and recall at the $nth$ threshold. 

 \begin{equation}
AP =  \sum_{n}(R_{n} -R_{n-1})P_{n}
\label{eq:2}
\end{equation}

AP metric is calculated at different IoU thresholds to get the best combination of IoU and AP for the object detection task.

\section{Methods}

In this section, we detail a data-driven deep learning method and a classical feature-based method to locate the table in images. We formulate the problem as a classic object detection problem, where the object of consideration will be a tabular structure present in the images. 

Deep learning-based object detection methods are proven to be very effective. We use RetinaNet \cite{retinanet} object detection method as baseline for this dataset.
RetinaNet achieved state-of-the-art performance on the COCO \cite{coco} dataset.  
We also use a selective search~\cite{ss} model for object detection to report baseline results using classical computer vision. 

In this section, we introduce the deep convolutional object detection method RetinaNet and the selective search-based objection detection method.

\subsection{RetinaNet for Table Detection}
RetinaNet is a one-stage, dense objective detection method designed using convolutional layers. It is consists of a backbone network and two task-specific heads. The backbone network is designed to model input image features using convolutional layers. The backbone network is a stack of multiple skip connection-based convolutional blocks. Task-specific heads are designed to get classification scores and the coordinates for each of the anchor bounding boxes. Figure~\ref{fig:method} shows a high-level depiction of the RetinaNet.
RetinaNet addresses the extreme foreground-background class imbalance problem observed in the object detector training by introducing a sample complexity-based weighted cross-entropy loss called focal loss. Focal loss enforces the trainer to focus on hard examples by assigning higher weights. 
Cross entropy loss for binary classification is defined as shown in the Eq~\ref{eq:ce}

 \begin{equation}
CE(p, y) = \left\{
\begin{array}{ll}
      -log(p) & if y = 1 \\
      -log(1-p) & otherwise\\
\end{array} 
\right. 
\label{eq:ce}
\end{equation}

\begin{table*}[h]
\begin{center}
  \begin{tabular}{ | p{3cm} | c |c|c|c| }
    \hline
    \thead{Datasets} & \thead{Samples} &\thead{Diversity} & \thead{Annotation Method} &  \thead{Modalities} \\ \hline
    ICDAR 2013 & 128 & 805.79 & Manual & Searchable PDFs \\ \hline
    TableBank & 417,234 & 755.32 & XML + Latex & Word and Latex documents \\ \hline
    Marmot & 2000 & 608.65 & Manual & Searchable PDFs \\ \hline
    UNLV & 427 & - & Manual & Scanned documents \\ \hline
    DeepFigures & 1.4M & - & XML + Latex & Word and Latex documents \\ \hline
    \textbf{STDW (ours)} & 7294 & \textbf{900.72} & Manual & Searchable PDFs and Scanned documents \\ \hline
  \end{tabular}
\end{center}
\caption{Comparisons between the STDW dataset and some of the other publicly available datasets for table detection. Our dataset provides the most diverse, high-quality samples.}
\label{table:comp_datasets}
\end{table*}

To handle the large class imbalance problem focal loss adds a modulating factor $(1-p_t)^\lambda$ to the original binary cross entropy loss function. Focal loss is defined as shown in the Eq~\ref{eq:focal1}, where $\lambda \geq 0$ is a tuneable parameter. 

 \begin{equation}
FL(p_t) = - (1 - p)^\lambda log(p_t)
\label{eq:focal1}
\end{equation}

In this work we use a $\alpha$-balanced variant of the focal loss as shown in the Eq~\ref{eq:focal2}

 \begin{equation}
FL(p_t) = - \alpha_t(1 - p)^\lambda log(p_t)
\label{eq:focal2}
\end{equation}

\subsection{Selective Search} 
Selective search is a method to detect all possible object bounding boxes in an input image. Selective search is combined with classical image-level features and a support vector machine to detect objects. It uses a hierarchical grouping algorithm to find all possible bounding boxes with an object, those bounding boxes can be overlapping. Selective search assigns objectness score to bounding boxes denoting the probability of object presence inside the bounding box. A bounding box with a high objectness score might contain the object of interest. 
For computing features of bounding boxes bag-of-words~\cite{bog1} with color-SIFT descriptors~\cite{sift} is used. For the classification of bounding boxes, a support vector machine with a histogram intersection kernel is used.

\subsection{Dataset Diversity}
We estimate diversity based on the dataset's sample variations in the latent space. The spread of the dataset in the latent space is defined as diversity. The larger the spread of the samples, the higher is the diversity. To calculate the latent representation of each of the samples we use an ImageNet~\cite{imagenet} pre-trained VGG-16~\cite{vgg} model's $block5\_pool$ layer. VGG-16 is a very deep convolutional model used for image-related problems. The latent representation is of dimension 25088 and computed on input image resized to $[224, 224, 3]$.  
Algorithm~\ref{algo:diversity} provides the full procedure for diversity computation using a deep convolutional encoder model.

\section{Experiments}
In our experiments, we evaluate the classical computer vision method and convolutional RetinaNet model on our dataset.

\begin{algorithm}
\caption{Dataset Diversity Metric}\label{euclid}
\begin{algorithmic}[1]

\State Initialize VGG-16 model with ImageNet weights.
\State $features \gets []$
\State // Compute samples latent representations
\For {$each \ image \ image\_filepath$} 
\State // Read, resize and normalize image
\State $image\_np = read\_image(image\_filepath)$
\State $image\_np = resize\_and\_normalize(image\_np)$
\State // Compute latent representation
\State $feature = VGG16(image\_np)$
\State $features \gets features \cup feature$
\EndFor
\State // Compute standard deviation across features dimension
\State $features\_std \gets compute\_std(features, axis=1)$
\State // Compute L2 norm
\State $diversity  = \lVert features\_std \rVert_2$

\end{algorithmic}
\label{algo:diversity}
\end{algorithm}

\begin{table}[h]
\begin{center}
  \begin{tabular}{ | p{3cm} | c |c|c| }
    \hline
    \thead{Method} & \thead{IOU} & \thead{AP} \\ \hline
    RetinaNet & 0.5 & 0.78 \\ \hline
    Selective Search  & 0.5 & 0.61 \\
    \hline
  \end{tabular}
\end{center}
\caption{Performance of classical and deep learning methods on STDW}
\label{table:stdw_perf}
\end{table}

\begin{table}[h]
\begin{center}
  \begin{tabular}{ | p{5cm} | c | }
    \hline
    \thead{Hyperparameter} & \thead{Value} \\ \hline 
    Base learning rate & 0.0004  \\  \hline
    Optimizer  & SGD  \\ \hline
    Momentum & 0.9 \\ \hline
    Batch size & 1 \\ \hline
    Buffer size & 20 * batch size \\ \hline
    Epochs & 30 \\ \hline
    Learning rate schedule & Constant \\ \hline
    Loss & Focal loss \\ \hline
    Feature extractor & Resnet-50 \\ \hline
    
    \hline
  \end{tabular}
\end{center}
\caption{Hyperparameters used for training the baseline RetinaNet model on STDW.}
\label{table:hp}
\end{table}

\begin{figure*}
  \centering
      \includegraphics[width=\textwidth, height=2.8in]{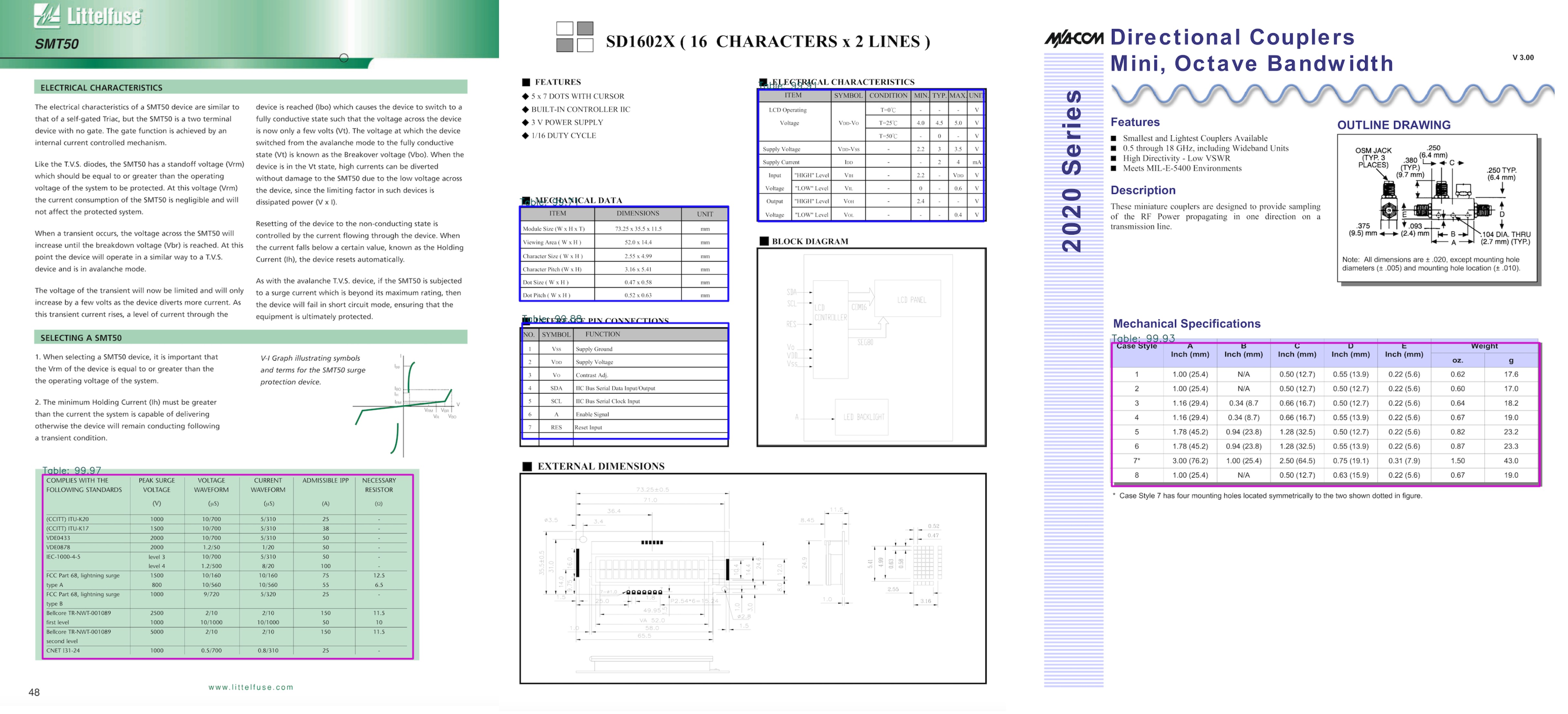}
\caption{Table Detection results using the baseline RetinaNet model on images from the Subex TDW daatset.}
\label{fig:result}
\end{figure*}

We have used Tensorflow~\cite{tf} and Datum~\cite{datum} for deep learning experiments described in this work. Tensorflow is used to build and train the RetinaNet model and Datum is used to prepare and load the dataset for training. 
Batch normalization\cite{bn} is used to reduce covariate shift and achieve faster convergence. Also, we have used the Nesterov momentum optimizer with a constant learning rate. RetinaNet model is trained using a GPU accelerator for 30 epochs to reach the baseline performance as reported in Table~\ref{table:stdw_perf}.
For selective search, we use the official Matlab code provided by the author. 

\subsection{Baseline results for STDW}
We provide baseline results for table detection on the STDW dataset using two methods. In one method we use the deep learning-based RetinaNet model and in the other method, we use a selective search-based object detection approach. For baseline results, IOU and AP are computed on the test set. We trained a RetinaNet model with XX feature extractor for detecting tables from input images. The hyperparameter set used in the RetinaNet experiment is shown in Table~\ref{table:hp}.
For a selective search-based approach we use a visual codebook of size 4000 and 4 levels spatial pyramid using a $1x1$, $2x2$, $3x3$ and $4x4$ division. This setting results in a total feature vector of length 360000.

\begin{figure}[h]
  \centering
      \includegraphics[width=3.6in,height=2.5in]{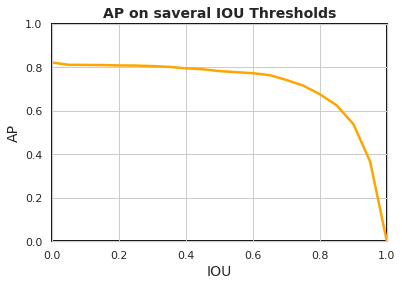}
\caption{IoU versus AP plot on the STDW test set using the RetinaNet baseline model.}
\label{fig:iou_vs_ap}
\end{figure}

Table~\ref{table:stdw_perf} shows the baseline results obtained using the deep learning-based RetinaNet model and classical selective search-based approach. RetinaNet performed significantly higher than that of the selective search method.
Figure~\ref{fig:iou_vs_ap} shows the variation of average precision with respect to intersection over the union of the predicted bounding boxes with that of the ground truth boxes. We report the AP metric value at the IoU threshold of 0.5. 

Results for a few images are visualized by drawing the predicted bounding boxes on the top of the original images as shown in Figure~\ref{fig:result}.  The baseline model predicted the locations of the tables correctly on the test images, though the predicted bounding boxes are not exactly precise, further investigation on the build building side is required to improve the results.  

Additionally, we have also computed the diversity metric for the STDW dataset and also for the TableBank, Marmot and ICDAR 2013 datasets. From Table~\ref{table:comp_datasets} diversity column we can see that the diversity of the proposed STDW dataset is more than the other open source datasets.

\section{Conclusion}
A large-scale table detection dataset is introduced in this paper. Out dataset includes 7K image samples collected from various open sources. Compared to the current datasets for this task, the STDW dataset is very diverse and contains high-quality manually annotated samples. 
The large scale and diverse nature of the data enable us to apply data-driven deep learning methods for table detection problems and achieve better performance. 
The provided baseline experimental results show the superiority of the data-driven deep learning approach over classical features-driven computer vision approaches.

\section*{Acknowledgement}
This research was carried out at Subex AI Labs. We gratefully acknowledge the support of team members in verifying the dataset annotation quality. 






%

\bibliographystyle{IEEEtran}
\bibliography{strings}

\end{document}